\documentclass{article}

\usepackage{arxiv}

\usepackage[utf8]{inputenc} 
\usepackage[T1]{fontenc}    
\usepackage{hyperref}       
\usepackage{url}            
\usepackage{booktabs}       
\usepackage{amsfonts}       
\usepackage{nicefrac}       
\usepackage{microtype}      
\usepackage{lipsum}		
\usepackage{graphicx}
\usepackage{natbib}
\usepackage{doi}

\title{Exploration of Algorithmic Trading Strategies for the Bitcoin Market}

\date{September, 2021}	

\author{\hspace{1mm}Nathan Crone \\
    	School of Computing\\
        Dublin City University\\
    	Dublin, Ireland \\
    	\texttt{nathan.crone2@mail.dcu.ie} \\
	\And
    	\hspace{1mm}Eoin Brophy \\
    	School of Computing and INFANT Research Centre\\
        Dublin City University\\
    	Dublin, Ireland \\
    	\texttt{eoin.brophy7@mail.dcu.ie} \\
	\And
    	\hspace{1mm}Tomás Ward \\
    	Insight SFI Centre for Data Analytics\\
        Dublin City University\\
    	Dublin, Ireland \\
    	\texttt{tomas.ward@dcu.ie} \\
}

\renewcommand{\shorttitle}{\textit{arXiv} Template}

\hypersetup{
pdftitle={A template for the arxiv style},
pdfsubject={q-bio.NC, q-bio.QM},
pdfauthor={David S.~Hippocampus, Elias D.~Striatum},
pdfkeywords={First keyword, Second keyword, More},
}

\begin{document}
\maketitle

\begin{abstract}
Bitcoin is firmly becoming a mainstream asset in our global society. Its highly volatile nature has traders and speculators flooding into the market to take advantage of its significant price swings in the hope of making money. This work brings an algorithmic trading approach to the Bitcoin market to exploit the variability in its price on a day-to-day basis through the classification of its direction. Building on previous work, in this paper, we utilise both features internal to the Bitcoin network and external features to inform the prediction of various machine learning models. As an empirical test of our models, we evaluate them using a real-world trading strategy on completely unseen data collected throughout the first quarter of 2021. Using only a binary predictor, at the end of our three-month trading period, our models showed an average profit of 86\%, matching the results of the more traditional buy-and-hold strategy. However, after incorporating a risk tolerance score into our trading strategy by utilising the model's prediction confidence scores, our models were 12.5\% more profitable than the simple buy-and-hold strategy. These results indicate the credible potential that machine learning models have in extracting profit from the Bitcoin market and act as a front-runner for further research into real-world Bitcoin trading.
\end{abstract}

\keywords{
          Time Series Classification \and
          Machine Learning \and
          Bitcoin \and
          Algorithmic Trading \and
          Blockchain
         }

\section{Introduction}

Cryptocurrencies have been one of the most popular assets to trade among retail investors over the last few years. This appeal has stemmed from their highly volatile nature, which has offered massive opportunities to make money. While human day traders have proved unsuccessful in predicting market fluctuations \citep{chague_day_2020}, algorithmic trading presents enormous potential in the space due to their high bandwidth and speed capabilities. Of all the cryptocurrencies which are hosted on exchanges, Bitcoin \citep{nakamoto_Bitcoin_2008} is by far the most valuable and most renowned, making up about 43\% of the entire cryptocurrency market. While an asset's value in and of itself does not directly translate into trading opportunity, Bitcoins market liquidity poses the perfect opportunity for algorithmic trading with 7\% of its total value changing hands daily \citep{Bitcoin_market_cap_2021}.

Trading algorithms work similarly to human traders wherein they draw upon their knowledge of the market's dynamics in combination with their perception of the current market state to make a judgement call on how to forecast the underlying asset's price. This research will focus on next-day Bitcoin price movement direction classification using machine learning algorithms. This one-day prediction period yielded favourable prediction results in previous studies \citep{mudassir_time_series_2020,sebastiao_forecasting_2021}, and is also a strong choice in terms of the availability of data measured on a daily timeframe. This is important as the features we give our algorithms dictate their ability to classify the daily price fluctuations.

Following on from the work of \citet{mudassir_time_series_2020}, we will use features describing the Bitcoin network alongside various technical indicator features based on these features. While features describing the internals of the Bitcoin network have shown to be successful \citep{balcilar_can_2017,ji_comparative_2019,huang_predicting_2019}, it is important to note that they lack other such market characteristics that some investors and speculators may consider to be useful. Some of these include stock market data, commodity data, currency exchange data, economic data, and social media data. Several studies have attempted to address this research gap on a small scale \citep{mai_how_2018,lyocsa_impact_2020,chen_Bitcoin_2020,mallqui_predicting_2019,jaquart_short_term_2021}; however, this study addresses feature expansion more comprehensively.

The key contributions of this paper can be summarised as follows:
\begin{itemize}
    \item We attempt to verify the performance of models reported in previous studies and demonstrate issues with respect to overfitting, something which is very common in published data science research.
    \item To the best of our knowledge, we establish a plausible new benchmark for machine learning in this space by representing the most rigorous and credible scientifically published Bitcoin trading algorithm to date.
    \item We develop an algorithmic Bitcoin trading method using a broad range of features which, when taken together, have been underexploited in previous studies. The use of these features in their raw form does not prove to have a significant effect on the model performance.
    \item We demonstrate the real-world trading performance of our developed model through an empirical test on completely unseen data collected during the year 2021. This provides a reliable metric for evaluating model performance over and above conventional metrics.
    \item We utilise the probabilistic outputs of the classifiers to naturally parameterise trading risk and allowing traders to flexibly specify their risk appetite. This shows traders with higher risk tolerances to be the most profitable.
\end{itemize}

We organise the rest of this paper as follows: Section \ref{reproducibility} investigates the reproducibility of a prominent paper in the area, Section \ref{methodologies} describes the approach we took to this forecasting problem, in Section \ref{results} we present the results of our machine-learning classification models hosting intermediate discussion between results, and Section \ref{conclusion} concludes our research.

\section{Reproducibility Challenges}
\label{reproducibility}

Our paper uses \citet{mudassir_time_series_2020} as a benchmark and attempts to reproduce its results. Mudassir et al. presented classification and regression machine learning approaches for predicting Bitcoin price movements. Their work uses internal Bitcoin features and technical indicators coupled with either Principal Component Analysis (PCA) or a novel feature selection technique to create a dataset for the modelling task. Their research proposes to build on previous works in the area, and as such, the paper uses models trained over three different time intervals to accurately compare the results with previous works. Due to the changing nature of the Bitcoin market, in this study, we will only consider the papers most recent interval, which runs from April 2013 to December 2019. The models presented in this study are an Artificial Neural Network (ANN), a Long-Short Term Memory Network (LSTM), a Stacked Artificial Neural Network (SANN), and a Support Vector Machine (SVM).

Using a single 80-20 train-test split, the paper reports results for both a next day price classification task, as shown in Table \ref{his_classification_results}, and a next day exact price regression task. The results of the classification task show the SANN to achieve the highest accuracy and AUC score, achieving 60\% for both metrics.

However, after implementing the methodology described in this work using the code on GitHub, we observed some discrepancies in the reported results. During this process, it became apparent that the SANN in both the classification task and regression task and the LSTM in the regression task had been trained on the test data. This oversight resulted in overfitting, thus the evaluation results derived for these models are excessively optimistic in terms of their predictional performance on future data. Table \ref{sann_table} highlights the significant impact of this overfitting on the classification task.

\begin{table}[h]
    \parbox{.45\linewidth}
    {
        \caption{The accuracy, f1-score, and AUC score reported in \citet{mudassir_time_series_2020} for their classification models.}
        \label{his_classification_results}
        \centering
        \begin{tabular}{llllll}
            \toprule
                Metrics   & ANN  & SVM  & SANN          & LSTM \\
            \midrule
                Acc. (\%) & 53   & 56   & \textbf{60}   & 54 \\
                F1-Score  & 0.61 & 0.53 & 0.60          & \textbf{0.66} \\
                AUC       & 0.53 & 0.56 & \textbf{0.60} & 0.54 \\
            \bottomrule
        \end{tabular}
    }
    \hfill
    \parbox{.45\linewidth}
    {
        \caption{A comparison of the SANN evaluation metrics reported in \citet{mudassir_time_series_2020} alongside metrics we derived after removing the model overfitting.}
        \label{sann_table}
        \begin{tabular}{llll}
            \toprule
                Metrics   & SANN (org) & SANN (fix)\\
            \midrule
                Acc. (\%) & 60         & 54 \\
                F1-score  & 0.60       & 0.51 \\
                AUC       & 0.60       & 0.54 \\
            \bottomrule
        \end{tabular}
    }
\end{table}

It also appears that another error occurs in the PCA step where the PCA components were not applied correctly. The models in this paper are trained using the principal components themselves as features instead of using these components to transform their data to then be used for model training. This produces erroneous features, completely distorting the data used in the PCA derived results.

Moreover, in the paper in question, hyperparameter tuning and experimentation with model architecture are not developed in good detail and there is no evidence to suggest that hyperparameter optimisation took place to the level of rigour required. Furthermore, some of the derived hyperparameter values are not well supported in the paper in terms of them being derived in a process-driven way. To improve the rigour of the approach described in the paper, it would have been better if hyperparameter optimisation for parameter selection was utilised and if the single 80-20 train-test split was replaced with cross-validation.

Our investigation of this study and examination of its work has highlighted the challenges facing data scientists in rigorous and robust model development. As the data science industry evolves, better approaches to evaluation are constantly being developed and in this example, we hope to progress this development by pointing out some common issues and how to remedy them. Despite these issues, we believe the authors of this paper are doing a great service by making their data and models freely and publicly available and it is thanks to the approach of these authors that we were able to highlight any issues with their process. While in the future others may find shortcomings with our approach, we hope that this ethos of openness and transparency of data science in the FinTech space will help to improve the robustness, trust, and utility of the underlying models and derived results.

\section{Methodologies}
\label{methodologies}

\begin{figure}[h]
\centering
\fbox{\includegraphics[width=\textwidth]{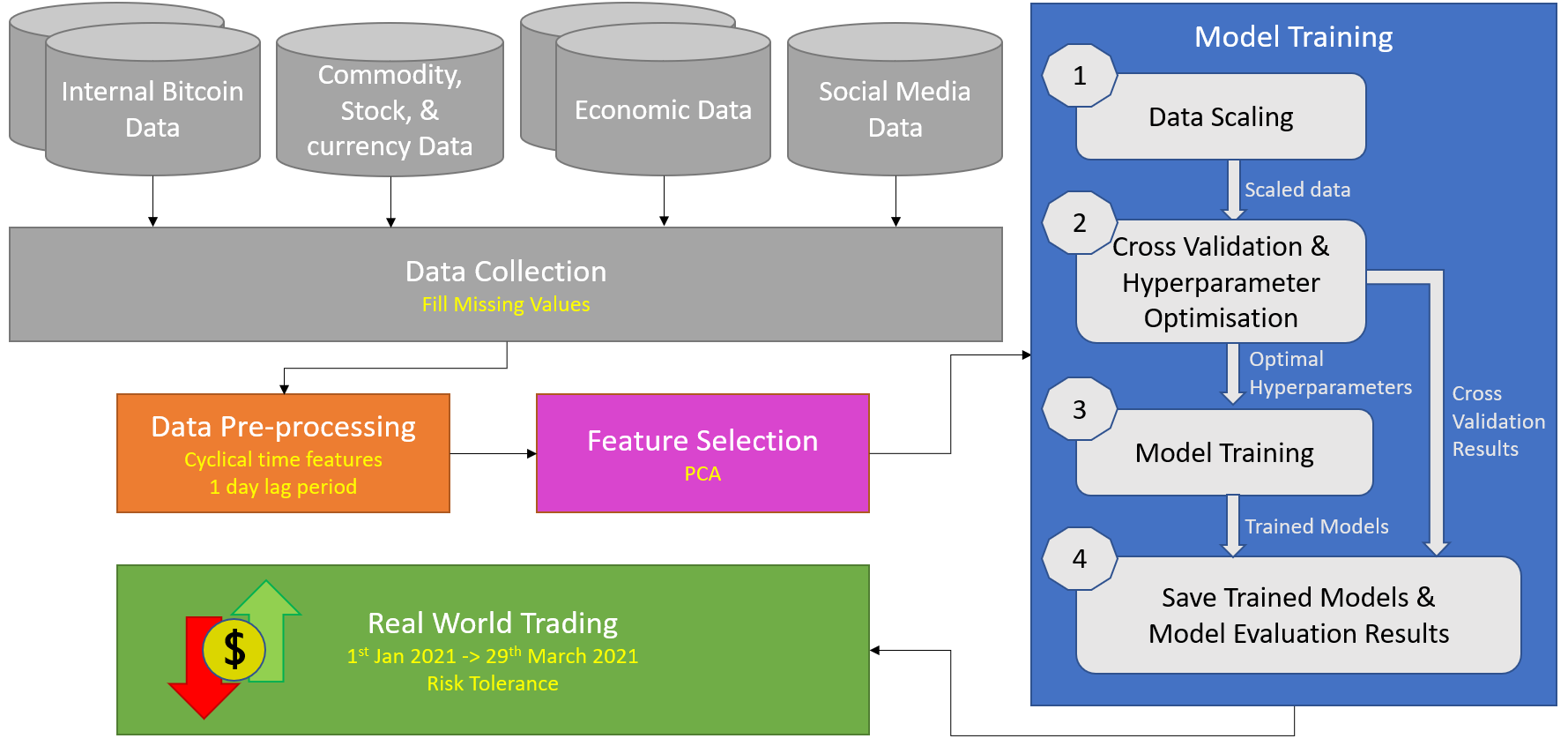}}
\caption{The proposed methodology for this study.}
\label{method_figure}
\end{figure}

Figure \ref{method_figure} shows the methodology proposed in this paper. The presented approach starts with the construction of a dataset using the features explained in Section \ref{data_collection}. We then use data pre-processing to gather the data, clean it, and create new aggregate features. In the feature selection step, we extract the relevant features from the data to use for model training. In training, we scale the data, define our parameters using hyperparameter optimisation alongside cross-validation, and train our models. The last step of our process involved employing our trained models in trading on real-world data as an intuitive form of evaluation.

\subsection{Data Collection}
\label{data_collection}

To build on previous research, we wanted to incorporate more features into the data than had been done in previous studies. These features break down into two distinct types: features internal to the Bitcoin Blockchain network and external factors that may affect the Bitcoin price.

Internal features have heavily informed research in this area due to their proven predictive power concerning the Bitcoin price. These features describe information related to the Bitcoin asset itself and the interactions on its Blockchain network. While many online sources contain this data, we decided to follow \citet{mudassir_time_series_2020} in scraping internal Bitcoin features from https://bitinfocharts.com. As explained in their paper, this website offers the ability to scrape nine different technical indicators over five different periods for each feature, alongside the raw feature values. These technical indicators allow us to capture underlying interdependencies and statistical factors in the data that may otherwise have gone unnoticed.
Additionally, we combine this data with other internal features scraped from https://data.Bitcoinity.org. This data source includes data from various Bitcoin exchanges throughout the world and adds several extra features to our feature set. Having data from multiple exchanges allows us to capture previously unknown variances in the data shown to Bitcoin traders on these platforms.

While previous studies have incorporated external features, such as a small set of stock or commodity indices \citep{mallqui_predicting_2019,chen_Bitcoin_2020} or macroeconomic and sentiment-based features \citep{mai_how_2018,lyocsa_impact_2020,raju_real-time_nodate}, these studies have not used these features simultaneously or on the scale proposed in this research. Alongside this, to the best of our knowledge, currency exchange data has never been explored as a potential predictive factor. As a novel contribution to our paper, we significantly expand the standard feature set used in this modelling task to explore untapped feature areas to increase our prediction power.

We scraped commodities, stocks, and currency exchange data from https://finance.yahoo.com using the Yahoo Finance API. This API provided us with a one-stop shop to scrape all the external features that we needed from these categories. Table \ref{yahoo_features} shows a list of the diverse set of commodity futures, stock exchange indices, and currency exchange rates that we incorporate into our model. Instead of focusing solely on monetary commodities like gold, as was done in earlier studies, we included a vast range of monetary, industrial, and food commodities in this study. This helps us to characterise market influencers like inflation which are sometimes misreported in economic indicators.
Furthermore, while previously cited studies only considered the leading American stock exchange indexes in their data, we take a more worldwide approach to features in this study. Bitcoin is an asset used and traded worldwide by people with vastly different market perceptions and using many different currencies. This must be factored into our data to allow our model to gain the best possible perception of the market that we can give it.

\begin{table}[ht]
    \begin{center}
        \caption{The commodity futures, stock indices, and currency exchange rates scraped from Yahoo Finance.}
        \label{yahoo_features}
        \resizebox{\columnwidth}{!}{%
            \begin{tabular}{lllll}
                \toprule
                    \multicolumn{2}{c}{Commodity Futures} & \multicolumn{2}{c}{Stock Indices} & \multicolumn{1}{c}{Currency Exchange Rates}\\
                \midrule
                    Crude Oil    & Soybean Oil & S\&P 500 Index               & Turkey's BIST 100 index        & Euro - GBP\\
                    Natural Gas  & Corn        & Dow Jones Industrial Average & Taiwan exchange index          & Euro - Swiss Franc\\
                    Gold         & Wheat       & Nasdaq index                 & Hong Kong's Hang Seng index    & Euro - Japanese Yen\\
                    Silver       & Oat         & NYSE composite index         & Singapore FTSE straits index   & GBP - Japanese Yen\\
                    Platinum     & Rough Rice  & AMEX composite index         & Japanese Nikkei 225 index      & USD - Japenese Yen\\
                    Palladium    & Sugar       & Russell 2000 index           & Korean Kospi index             & USD - Euro\\
                    Copper       & Cocoa       & Euro STOXX 50 index          & Indonesian IDX composite index & USD - Canadian Dollar\\
                    Aluminium    & Coffee      & Euronext 100 index           & Australian ASX 200 index       & USD - Australian Dollar\\
                    Lumber       & Live Cattle & UK FTSE 100 index            & Australian ordinaries index    & USD - Mexican Peso\\
                    Cotton       & Lean Hogs   & Irish ISEQ index             & Johannesburg top 40 index      & USD - Hong Kong Dollar\\
                    Soybean Meal &             & German DAX index             & Buenos Aires S\&P MERVAL index & \\
                                 &             & Belgium 20 index             & Santiage IPSA index            & \\
                                 &             & French CAC index             & Mexican MXX index              & \\
                                 &             & Spanish IBEX index           & Toronto's S\&P/TSX index       & \\
                \bottomrule
            \end{tabular}
        }
    \end{center}
\end{table}

Globally diverse economic data was another crucial factor of our novel feature set. Economic influences have become particularly important over the last few years and months, where we have seen massive monetary expenditure and changes in fiscal policy. We scraped the most critical US Federal Reserve Bank (FED) economic indicators from http://quandl.com using their API, and we got access to vital data on other EU and non-EU countries using https://db.nomics.world/Eurostat.

The final set of features built into our dataset relate to social media sentiment. Over the last number of months, social media has had a powerful influence on the trading landscape of specific assets \citep{lyocsa_yolo_2021}. Our study attempts to capture some of these social media influences by paying particular attention to the Bitcoin-related Twitter activity. Through https://bitinfocharts.com, we were able to scrape a counter of the Bitcoin-related tweets for each day, alongside several technical indicator features based on this. We could then add to this by using the Twitter API to scrape tweets related to specific Bitcoin influencers. Our logic for this was influenced by \citet{mai_how_2018} as they report the majority of Bitcoin-related tweets to be noise and that some minority of users and tweets have a significant impact on the price. Due to the restrictions of the Twitter API limit, we could only get access to the tweets of Elon Musk. Musk's tweets have had a noticeable impact on the Bitcoin price in recent months, and by transforming his tweets into sentiment-based numeric features, we hope to represent this influence in our dataset.

\subsection{Data Pre-Processing}

In data pre-processing, we fill in the missing values from the collected data. Where possible, we imputed missing values in the internal Bitcoin data using linear interpolation. Missing values that cannot be imputed are filtered out later through the date range selected for the training interval. In \citet{mudassir_time_series_2020}, they fill these uninterpolatable missing values with the most common value for that column; however, we did not feel that this approach is appropriate given the data's sequential nature and Bitcoin's growth over the past decade. The commodity, stock, and currency exchange data had missing values that needed to be filled from when their respective exchanges were closed. We use forward filling for the price features and handle the volume feature by filling them with ‘0'. This means that the price value from the previous trading day is carried forward and used as the current day's value. We also used this technique to populate readings in the economic indicator data. Interpolation was not an option in this case as, in the real world, we do not know what the next value will be, so we can only base our market perceptions on the previous value. To help account for this by giving our algorithm a better sense of time, we included two cyclical time features to represent the day of the week and the day of the month.

In addition to this, we created a period lag feature as in \citet{laboissiere_maximum_2015} to represent our classification target. For the lag period of one day used in this study, this created a feature that would be assigned a '1' or a '0' based on whether the current price had increased or decreased relative to the previous day's price.

\subsection{Feature Selection}

Given that our feature expansion left us with over 1,450 features, feature selection was vital in the modelling process. The approach we took for this was dimensionality reduction through Principal Component Analysis (PCA). This data projection technique uses matrix manipulations to define a set of components that can transform our features into a new set of linearly independent features. Using this feature selection method, we could capture different amounts of variance in our data through a much smaller feature set than our original dataset.

In this step, we also select the date range for our training data. We employ a data-driven approach when making this selection by attempting to maximise the interval size while ensuring not to bring missing values into the data. The resulting range for our training data was from August 2015 to December 2020, leaving us a quarter from January 2021 to April 2021 to trade with our models on unseen real-world data.

\subsection{Prediction Models}

\begin{figure}[h]
    \centering
    \fbox{\includegraphics[width=\textwidth]{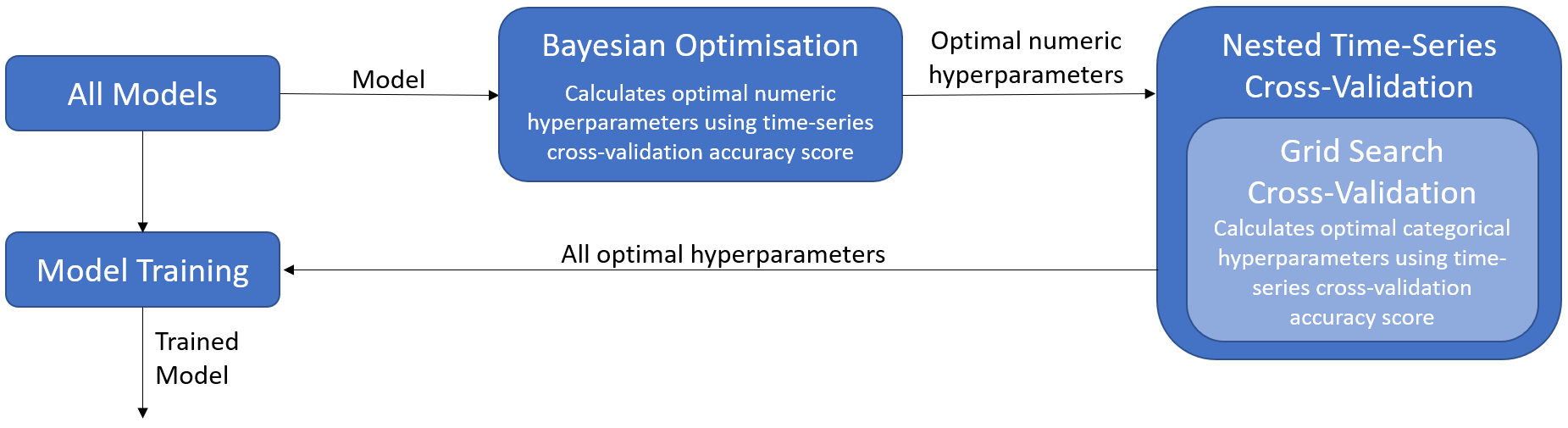}}
    \caption{The hyperparameter optimisation process proposed in this study.}
    \label{hyper_opt_and_cv}
\end{figure}

Our training data consists of only 4500 data points, one entry for each day; hence, we felt that it was not sufficiently large enough to train a neural-based model. Instead, in this paper, we explore simpler models such as Support Vector Machine (SVM), XGBoost (XGB), Random Forest Classifier (RFC), and Bernoulli Naive Bayes (BNB). To ensure we were not overfitting, we rigorously use nested time-series cross-validation during our hyperparameter optimisation step, making use of the sequential nature of the time-series throughout its folds. This process also provides us with a realistic evaluation metric for our model. Figure \ref{hyper_opt_and_cv} shows the incorporation of this cross-validation into the hyperparameter optimisation process. Due to computational limitations, we split the hyperparameter optimisation into two steps, dealing with the numeric values through Bayesian Optimisation and utilising grid search cross-validation to select the categorical values. Once we select the optimal hyperparameters and calculate the cross-validation evaluation metrics, we use the chosen hyperparameters to train our final prediction models on the full dataset. This final model yields the data's absolute predictive power, which sets up nicely for our evaluation through real-world trading.

\subsection{Trading}

We trade with these algorithms over completely unseen data collected throughout 2021 by using each model to make a following day price direction classification. Regardless of the asset price on any day, under our trading strategy, if the model predicts a price increase, we buy one Bitcoin, and if the model predicts a price decrease, we sell one Bitcoin. The following day, we then neutralise our position by either selling any Bitcoin we purchased or buying back any Bitcoins we sold. This daily refresh allows us to continually update a total profit \& loss tracker for our models, giving us a straightforward way to measure the real-world predictive power of our models. We define this metric as the difference between our buy price and our sell price, regardless of the order they occur. We can also take our trading strategy further by incorporating a risk tolerance score that employs the model's confidence scores in the underlying predictions. Through normalisation, we parameterise the fluctuations in model confidence through a customisable risk tolerance parameter, introducing the ability for a trader to sit out of the market during periods of model uncertainty.

\section{Results \& Discussion}
\label{results}

\subsection{Price Direction Forecast Accuracy}

We initially evaluated the performance of our classification models through nested cross-validation by recording the cross-validation accuracy, precision, recall, and F1 of the best model configurations. The most desirable model maximises all of these metrics; however, accuracy is the most commonly used, and hence we use it to optimise our hyperparameters. Table \ref{my_results_table} puts forward our realistic and thoroughly validated results for the predictability of Bitcoin's price movements over our model training period. We derive these results from three datasets, each obtained using a specific number of PCA components, to explain various levels of variance of the original data. The evaluation of these models across these datasets showed the SVM to perform the best, achieving an accuracy score of 56\% and an F1 score of 0.716 when trained on the dataset explaining 95\% of the original data's variance. These results were closely followed by that of the RFC, which achieved an average cross-validation accuracy score of 55.7\% on this same dataset.

\begin{table}[h]
    \caption{The cross-validation evaluation metrics for each model after being trained on datasets maintaining different levels of explained variance from the original data.}
    \label{my_results_table}
    \centering
    \begin{tabular}{llllll}
        \toprule
            Metrics    & Expl var (\%)   & SVM            & XGB            & RFC \\
        \midrule
                       & 80              & 55.9           & 54.7          & 54 \\
            Acc. (\%)  & 90              & 54.9           & 55           & 55.7 \\
                       & 95              & \textbf{56}    & 53.9          & 55.7 \\
        \midrule
                       & 80              & \textbf{0.716} & 0.654          & 0.687 \\
            F1-Score   & 90              & 0.69           & 0.688          & 0.707 \\
                       & 95              & \textbf{0.716} & 0.441          & 0.704 \\
        \midrule
                       & 80              & 0.56           & 0.56           & 0.556 \\
            Precision  & 90              & 0.552          & 0.56           & 0.56 \\
                       & 95              & 0.56           & 0.349          & \textbf{0.562} \\
        \midrule
                       & 80              & \textbf{1}     & 0.833          & 0.911 \\
            Recall     & 90              & 0.938          & 0.907          & 0.964 \\
                       & 95              & \textbf{1}     & 0.6            & 0.947 \\
        \bottomrule
    \end{tabular}
\end{table}

To further ensure the rigour of our process, we ran AutoML on our data as a sanity check of our derived results. Under the AutoML process, our data is loaded in, scaled, the key features are selected, and then the models in the AutoML pipeline are iteratively evaluated to see which one scores the best. The process stops once it finds the optimal model, and then it returns the necessary code to create this model and its associated cross-validation accuracy score. The model selected through our run was the Bernoulli Naive Bayes (BNB) model, which achieved an average cross-validation accuracy score of 57\%. This accuracy score was remarkably similar to the scores we got through our hyperparameter optimisation and cross-validation process and hence backed up the validity of our results.

\subsection{Feature Significance}

After defining our methodology and obtaining our evaluation scores, we then set out to measure the difference in performance between our features and the features set out in \citet{mudassir_time_series_2020}. We saw no significant difference in the reported evaluation metrics upon replacing our features with the features used in this study. This was surprising given that we had almost doubled the feature set size; however, when we investigated the feature importance in more detail, the significance of the technical indicators used in conjunction with the features in the previous study became more apparent. Using a random forest classifier, we calculated an importance score for each feature in our dataset and discovered that of the top 200 most important features, only 20\% were added to the feature set in this study, of which 78\% were technical indicators. This analysis emphasises the predictive power of technical indicators and sets the stage for future research pairing similar features with relevant technical indicators.

\subsection{Trading Performance}

After evaluating our models using our defined trading strategy over the three-month real-world test period, our models made an average of \$24,000. While we expected our models to actively trade the market using market fluctuations, from analysis of the prediction scores made by these models, they seem to opt for more of a buy-and-hold strategy to trading the Bitcoin market. This strategy proved massively successful over the given period, with the Bitcoin price starting at over \$28,000 and rising above \$52,000 by the end, constituting an 86\% increase in price.

\begin{figure}[h]
    \centering
    \fbox{\includegraphics[width=\textwidth]{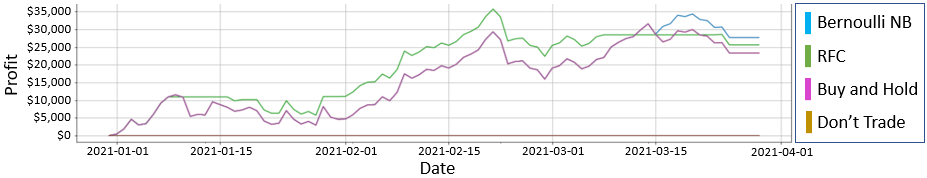}}
    \caption{The models calculated profit and loss over time relative to the Bitcoin price as we trade with a 30\% risk tolerance using its predictions.}
    \label{trading_figure}
\end{figure}

Despite the overwhelming majority of the models predicting the naive class each time, as we varied our risk tolerance parameter, we saw more dissimilarities between the models trading profit \& loss values. While in a real-world situation, this parameter would be set based on the traders personal trading goals, for this study, we wanted to investigate what risk tolerance might yield the most profit. This investigations stems from the fact that the lower a traders risk tolerance is, the more hesitant they will be to either buy or sell at any given time. Wielding the probabilistic prediction outputs of our models, we created a ROC curve and used its geometric mean scores to obtain the optimal classification threshold for each model. We could then translate these thresholds into a risk tolerance parameter through normalisation. Figure \ref{trading_figure} shows the profit \& loss values for a few of our models over the given period when we specified a 30\% risk tolerance. This figure illustrates that both the Bernoulli Naive Bayes (BNB) model and the Random Forest Classifier outperform a buy-and-hold trading strategy, with the BNB model achieving a profit of over \$27,000 while the buy-and-hold strategy only saw a profit of \$24,000. This difference represents a relative profit of 12.5\% over the quarter for the BNB model. While this optimised risk parameter yielded desirable results, most risk tolerance values could not outperform this simple buy-and-hold strategy.

\subsection{Limitations}
Results different to those presented in this paper may be obtained when the models are run using our code from GitHub. This stems from the fact that the derived results presented here were obtained during the period of 23rd March to 2nd April, after which some changes were made concerning the data used to train the model among other minor procedural changes. As such, the results generated in this paper are based on a subset of the data collected through our scraping process and using a very similar process to that of our current Jupyter notebooks.

\section{Conclusion \& Future Work}
\label{conclusion}

This paper sets out to clear up much of the existing literature by proposing a new benchmark bitcoin price forecasting process that future studies can build on. We explore the reproducibility challenges currently facing the data science industry and demonstrate the need for a rigorous machine learning pipeline to ensure the accurate reporting of model results. Alongside this, we propose a new, more realistic form of evaluation through the use of real-world trading, and we investigate the effect a trader's risk tolerance has on their profitability. This showed our models to trade more similar to a buy-and-hold strategy than what would be expected of an algorithmic trading algorithm. Nevertheless, this proved hugely successful over our evaluation period.
We were also able to exhibit the capabilities of building a modifiable risk tolerance parameter into future trading algorithms through our Bernoulli Naive Bayes model, which, relative to a buy-and-hold strategy, achieved a 12.5\% better quarterly return.

While our vast expansion of previously defined feature sets did not significantly impact our models forecasting ability, future inquiry should attempt to build on this by further exploring the use of technical indicators on these. Moreover, we feel further exploration into social media such as Reddit, Facebook, 4chan, and Twitter would boost model forecasting ability.
In addition, online learning could be applied to the trading evaluation to increase its validity as a metric that can quantify a models profitability in the real-world bitcoin market.

We appreciate that the approach we present in this paper is unlikely to be without error and in the spirit of openness and transparency, we make our code and data available through our GitHub repository (https://github.com/Crone1/Bitcoin-Algorithmic-Trading-Paper) to allow full replication by others. We welcome any extensions, corrections, and improvements to the models we present.

\bibliographystyle{unsrtnat}
\bibliography{algo_trading} 

\begin{thebibliography}{16}
\providecommand{\natexlab}[1]{#1}
\providecommand{\url}[1]{\texttt{#1}}
\expandafter\ifx\csname urlstyle\endcsname\relax
  \providecommand{\doi}[1]{doi: #1}\else
  \providecommand{\doi}{doi: \begingroup \urlstyle{rm}\Url}\fi

\bibitem[Chague et~al.(2020)Chague, De-Losso, and Giovannetti]{chague_day_2020}
Fernando Chague, Rodrigo De-Losso, and Bruno Giovannetti.
\newblock Day {Trading} for a {Living}?
\newblock {SSRN} {Scholarly} {Paper} ID 3423101, Social Science Research
  Network, Rochester, NY, June 2020.
\newblock URL \url{https://papers.ssrn.com/abstract=3423101}.

\bibitem[Nakamoto(2008)]{nakamoto_Bitcoin_2008}
Satoshi Nakamoto.
\newblock Bitcoin: {A} {Peer}-to-{Peer} {Electronic} {Cash} {System}.
\newblock \emph{Cryptography Mailing list at https://metzdowd.com}, page~9,
  October 2008.

\bibitem[CoinMarketCap(2021)]{Bitcoin_market_cap_2021}
CoinMarketCap.
\newblock Market capitalization and trading volume of bitcoin from march 2013
  to may 2021, 2021.
\newblock URL \url{https://coinmarketcap.com/currencies/bitcoin/}.

\bibitem[Mudassir et~al.(2020)Mudassir, Bennbaia, Unal, and
  Hammoudeh]{mudassir_time_series_2020}
Mohammed Mudassir, Shada Bennbaia, Devrim Unal, and Mohammad Hammoudeh.
\newblock Time-series forecasting of {Bitcoin} prices using high-dimensional
  features: a machine learning approach.
\newblock \emph{Neural Computing and Applications}, July 2020.
\newblock ISSN 1433-3058.
\newblock \doi{10.1007/s00521-020-05129-6}.
\newblock URL \url{https://doi.org/10.1007/s00521-020-05129-6}.

\bibitem[Sebastião and Godinho(2021)]{sebastiao_forecasting_2021}
Helder Sebastião and Pedro Godinho.
\newblock Forecasting and trading cryptocurrencies with machine learning under
  changing market conditions.
\newblock \emph{Financial Innovation}, 7\penalty0 (1):\penalty0 3, January
  2021.
\newblock ISSN 2199-4730.
\newblock \doi{10.1186/s40854-020-00217-x}.
\newblock URL \url{https://doi.org/10.1186/s40854-020-00217-x}.

\bibitem[Balcilar et~al.(2017)Balcilar, Bouri, Gupta, and
  Roubaud]{balcilar_can_2017}
Mehmet Balcilar, Elie Bouri, Rangan Gupta, and David Roubaud.
\newblock Can volume predict {Bitcoin} returns and volatility? {A}
  quantiles-based approach.
\newblock \emph{Economic Modelling}, 64:\penalty0 74--81, August 2017.
\newblock ISSN 0264-9993.
\newblock \doi{10.1016/j.econmod.2017.03.019}.
\newblock URL
  \url{http://www.sciencedirect.com/science/article/pii/S0264999317304558}.

\bibitem[Ji et~al.(2019)Ji, Kim, and Im]{ji_comparative_2019}
Suhwan Ji, Jongmin Kim, and Hyeonseung Im.
\newblock A {Comparative} {Study} of {Bitcoin} {Price} {Prediction} {Using}
  {Deep} {Learning}.
\newblock \emph{Mathematics}, 7\penalty0 (10):\penalty0 898, October 2019.
\newblock \doi{10.3390/math7100898}.
\newblock URL \url{https://www.mdpi.com/2227-7390/7/10/898}.
\newblock Number: 10 Publisher: Multidisciplinary Digital Publishing Institute.

\bibitem[Huang et~al.(2019)Huang, Huang, and Ni]{huang_predicting_2019}
Jing-Zhi Huang, William Huang, and Jun Ni.
\newblock Predicting bitcoin returns using high-dimensional technical
  indicators.
\newblock \emph{The Journal of Finance and Data Science}, 5\penalty0
  (3):\penalty0 140--155, September 2019.
\newblock ISSN 2405-9188.
\newblock \doi{10.1016/j.jfds.2018.10.001}.
\newblock URL
  \url{http://www.sciencedirect.com/science/article/pii/S2405918818300928}.

\bibitem[Mai et~al.(2018)Mai, Shan, Bai, Wang, and Chiang]{mai_how_2018}
Feng Mai, Zhe Shan, Qing Bai, Xin~(Shane) Wang, and Roger H.~L. Chiang.
\newblock How {Does} {Social} {Media} {Impact} {Bitcoin} {Value}? {A} {Test} of
  the {Silent} {Majority} {Hypothesis}.
\newblock \emph{Journal of Management Information Systems}, 35\penalty0
  (1):\penalty0 19--52, January 2018.
\newblock ISSN 0742-1222.
\newblock \doi{10.1080/07421222.2018.1440774}.
\newblock URL \url{https://doi.org/10.1080/07421222.2018.1440774}.
\newblock Publisher: Routledge \_eprint:
  https://doi.org/10.1080/07421222.2018.1440774.

\bibitem[Lyócsa et~al.(2020)Lyócsa, Molnár, Plíhal, and
  Širaňová]{lyocsa_impact_2020}
Štefan Lyócsa, Peter Molnár, Tomáš Plíhal, and Mária Širaňová.
\newblock Impact of macroeconomic news, regulation and hacking exchange markets
  on the volatility of bitcoin.
\newblock \emph{Journal of Economic Dynamics and Control}, 119:\penalty0
  103980, October 2020.
\newblock ISSN 0165-1889.
\newblock \doi{10.1016/j.jedc.2020.103980}.
\newblock URL
  \url{http://www.sciencedirect.com/science/article/pii/S0165188920301482}.

\bibitem[Chen et~al.(2020)Chen, Li, and Sun]{chen_Bitcoin_2020}
Zheshi Chen, Chunhong Li, and Wenjun Sun.
\newblock Bitcoin price prediction using machine learning: {An} approach to
  sample dimension engineering.
\newblock \emph{Journal of Computational and Applied Mathematics},
  365:\penalty0 112395, February 2020.
\newblock ISSN 0377-0427.
\newblock \doi{10.1016/j.cam.2019.112395}.
\newblock URL
  \url{http://www.sciencedirect.com/science/article/pii/S037704271930398X}.

\bibitem[Mallqui and Fernandes(2019)]{mallqui_predicting_2019}
Dennys C.~A. Mallqui and Ricardo A.~S. Fernandes.
\newblock Predicting the direction, maximum, minimum and closing prices of
  daily {Bitcoin} exchange rate using machine learning techniques.
\newblock \emph{Applied Soft Computing}, 75:\penalty0 596--606, February 2019.
\newblock ISSN 1568-4946.
\newblock \doi{10.1016/j.asoc.2018.11.038}.
\newblock URL
  \url{http://www.sciencedirect.com/science/article/pii/S1568494618306707}.

\bibitem[Jaquart et~al.(2021)Jaquart, Dann, and
  Weinhardt]{jaquart_short_term_2021}
Patrick Jaquart, David Dann, and Christof Weinhardt.
\newblock Short-term bitcoin market prediction via machine learning.
\newblock \emph{The Journal of Finance and Data Science}, 7:\penalty0 45--66,
  November 2021.
\newblock ISSN 2405-9188.
\newblock \doi{10.1016/j.jfds.2021.03.001}.
\newblock URL
  \url{https://www.sciencedirect.com/science/article/pii/S2405918821000027}.

\bibitem[Raju and Tarif(2020)]{raju_real-time_nodate}
S~M Raju and Ali~Mohammad Tarif.
\newblock Real-time prediction of bitcoin price using machine learning
  techniques and public sentiment analysis, 2020.

\bibitem[\v{S}tefan Ly\'{o}csa et~al.(2021)\v{S}tefan Ly\'{o}csa, Baum\"{o}hl,
  and V\^{y}rost]{lyocsa_yolo_2021}
\v{S}tefan Ly\'{o}csa, Eduard Baum\"{o}hl, and Tom\'{a}\v{s} V\^{y}rost.
\newblock Yolo trading: Riding with the herd during the gamestop episode.
\newblock Technical report, Leibniz Information Centre for Economics, Kiel,
  Hamburg, Kiel, Hamburg, 2021.
\newblock URL \url{http://hdl.handle.net/10419/230679}.

\bibitem[Laboissiere et~al.(2015)Laboissiere, Fernandes, and
  Lage]{laboissiere_maximum_2015}
Leonel~A. Laboissiere, Ricardo A.~S. Fernandes, and Guilherme~G. Lage.
\newblock Maximum and minimum stock price forecasting of {Brazilian} power
  distribution companies based on artificial neural networks.
\newblock \emph{Applied Soft Computing}, 35:\penalty0 66--74, October 2015.
\newblock ISSN 1568-4946.
\newblock \doi{10.1016/j.asoc.2015.06.005}.
\newblock URL
  \url{https://www.sciencedirect.com/science/article/pii/S156849461500352X}.

\end{thebibliography}

\end{document}